\pdfoutput=1

\documentclass[11pt]{article}

\usepackage{shortcuts}

\title{Algorithm for Automatic Legislative Text Consolidation}

\author{Matias Etcheverry \\
  Doctrine \\
  Ecole Nationale Supérieure \\ Paris-Saclay \\
  \href{mailto:matias.etcheverry@doctrine.fr}{matias.etcheverry@doctrine.fr} \\\And
  Thibaud Real\\
  Doctrine  \\
  \href{mailto:thibaud.real-del-sarte@doctrine.fr}{thibaud.real-del-sarte@doctrine.fr}\
  \And
  Pauline Chavallard\\
  Doctrine  \\
  \href{mailto:pauline@doctrine.fr}{pauline@doctrine.fr}}

\begin{document}

\maketitle

\begin{abstract}
  This study introduces a method for automating the consolidation process in a legal context, a time-consuming task traditionally performed by legal professionals. We present a generative approach that processes legislative texts to automatically apply amendments. Our method employs light quantized generative model, finetuned with LoRA, to generate accurate and reliable amended texts. To the authors knowledge, this is the first time generative models are used on legislative text consolidation. Our dataset is publicly available on HuggingFace\footnote{\href{https://huggingface.co/datasets/DoctrineAI/legal_consolidation}{Link to dataset}}. Experimental results demonstrate a significant improvement in efficiency, offering faster updates to legal documents. A full automated pipeline of legislative text consolidation can be done in a few hours, with a success rate of more than 63\% on a difficult bill.
\end{abstract}

\section{Introduction}
Every year in France, the \textit{Projet de Loi Finance}\footnote{\href{https://www.assemblee-nationale.fr/dyn/opendata/PRJLANR5L16BTA0223.html}{Link to \textit{Projet de Loi Finance} for 2024}} (PLF), annually introduces numerous modifications to the General Tax Code (484 in 2024). The objective of this study is to automate the process of legislative text consolidation, which is the act of combining modifications from a \texttt{modification section}, contained inside the PLF, to an \texttt{existing article} to generate a \texttt{modified article}.  Example \ref{triplet:dummy_example} illustrates a dummy consolidation, where the original text of a law is updated by incorporating amendments directly into it, resulting in a revised, coherent version.

\begin{triplet}[Illustration of legislative consolidation]
    \label{triplet:dummy_example}\artquote{Existing article}{Paris is the capital of France.}

    \artquote{Modification section}{\newline
        I.- Replace the word « is » with « has been ».\newline
        II.- Add « since the late 10th century » at the end of the sentence.\newline}
    \artquote{Modified article}{Paris has been the capital of France since the late 10th century.}
\end{triplet}

Legislative text consolidation is a critical yet time-consuming task, traditionally performed manually by legal professionals. A sample of the PLF is presented in Example\footnote{All examples are translated from French to English.} \ref{triplet:introduction_triplet}. It modifies \texttt{article 1586 ter} and \texttt{article 1586 quater} of the General Tax Code.

\begin{triplet}[Extract of article 79 of the PLF 2024]
    \label{triplet:introduction_triplet}
    \artquote{Modification sections}{\newline
        I.-The General Tax Code is amended as follows:\newline
        A.-The following words are added to the first sentence of the second paragraph of 1 of II of Article 1586 ter: « , as it stood prior to Finance Act 2023-1322 of 29 December 2023 for 2024 »;\newline
        B.-Article 1586 quater is amended as follows:\newline
        1° I is amended as follows\newline
        a) The second paragraph of b and c is amended as follows:\newline
        -at the beginning, the rate: « 0.125\% » is replaced by the rate: « 0.094\% »;\newline
        -at the beginning, the rate: « 0.094\% » is replaced by the rate: « 0.063\% »;\newline
        b) The second paragraph of c is amended as follows:\newline
        -the rate: « 0.225\% » is replaced by the rate: « 0.169\% »;\newline
        -the rate: « 0.113\% » is replaced by the rate: « 0.056\% »;}


\end{triplet}

\begin{figure}[ht]
    \centering
    \includegraphics[width=\columnwidth]{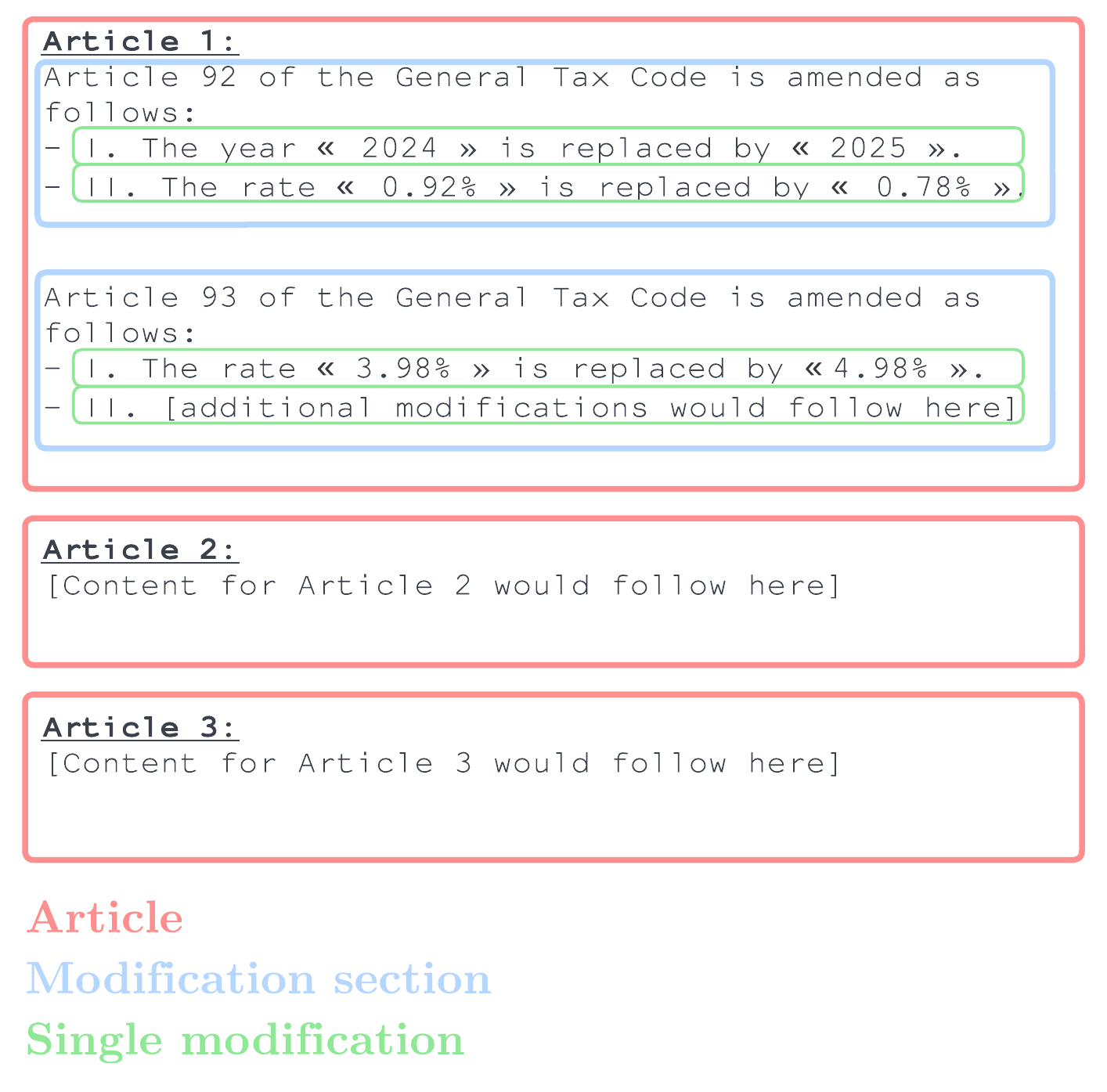}
    \caption{General structure of the PLF}
    \label{fig:structure_of_the_plf}
\end{figure}

The conventions represented on Figure~\ref{fig:structure_of_the_plf} are adhered to:
\begin{itemize}
    \item A legislative bill is composed of multiple articles\footnote{An article would be the equivalent of a section in a bill in common-law countries.}.
    \item An article comprises several sections. A section is defined as a collection of paragraphs that enact modifications to a single article.
    \item A section may effectuate either a singular modification or multiple modifications. For instance, section {\quotefont{A.-}} implements a single modification, whereas section {\quotefont{B.-}} introduces four modifications.
\end{itemize}

Three primary modification categories are identified: a deletion, involving the removal of a word, sentence, or paragraph; an addition, encompassing the insertion of a word, sentence, or paragraph and a substitution, where a word, sentence, or paragraph is exchanged for another. Example \ref{triplet:introduction_triplet} demonstrates one instance of addition and four instances of substitutions.

The automation of legislative text consolidation has the potential to significantly expedite this process, offering a rapid update of legal documents post-enactment and potentially pre-enactment, thereby enhancing the accessibility and reliability of legal information.

\section{Related works}
\subsection{Information extraction approaches}
\label{sec:information_extraction}

Modification sections typically follow a consistent lexical structure. \citet{moore95tree, moore97tree} and \citet{Mazzei_classification} exploit this formal consistency to extract amendments and construct a structured tree representation, applying information extraction techniques.

Subsequently, a clear trend is drawn in information extraction between tagging-based methods and generative methods. Tagging-based methods are designed to classify individual tokens (token-based methods) or clusters of tokens (span-based methods). In contrast, generative methods are oriented towards producing textual content that is inherently construed as a relationship triplet. Hence, \citet{shi2019simple} undertake a notably question answering adaptation of the BERT model to facilitate generation across a diverse corpus, achieving a remarkably good baseline. In more recent times, generative models appear to exhibit superior performance. \citet{josifoski2022genie} introduce the GenIE model, which succeeds in generating generation triplets through its utilization of the BART architecture.

In recent developments, models dedicated to text editing have garnered interest for their utility in tasks that necessitate the rearrangement of words and text spans, such as summarization. \citet{malmi2019encode} introduced \texttt{LaserTagger}, an approach that assigns one of several tags to tokens, including \texttt{KEEP}, \texttt{DELETE}, \texttt{SWAP}, or \texttt{PRONOMINALIZE}, to facilitate text editing. Concurrently, \citet{mallinson2020felix} developed a two-stage algorithm wherein the first model tags tokens, and the subsequent model is responsible for the rearrangement of these tagged tokens.

\subsection{Generative approaches}

Generative approaches rapidly took the lead to reinterpret any task of extraction, classification, or edition as generative problems under certain frameworks~\cite{t5}. Building upon this foundational work, \citet{flanT5} expanded the utility of these models through the fine-tuning process to accommodate a broad spectrum of human instructions, thereby enhancing their applicability. This advancement has catalyzed subsequent research endeavors, focusing extensively on the exploration of instruction-based fine-tuning within the realm of generative models.

\paragraph{Instruction tuning} It is crucial to recognize that fine-tuning the model for a specific task is pivotal \citep{gpt3, flan}. In specific-use Pretrained Large Language Models (PLLMs), such as for legislative text consolidation, we may use instruction tuning to ensure that our model consolidates the provided legal text in all cases.

\paragraph{Finetuning} Existing parameter-efficient tuning methods still lag behind full fine-tuning on higher-resource and challenging tasks, but often succeed when dealing simple tasks, as consolidation would be \citep{he2022unified}. These approaches enable instruction tuning to be performed on cost-effective GPUs.

On one hand, prompt tuning methods involve concatenating the embeddings of input tokens with a trainable tensor. This tensor can be optimized through backpropagation to enhance the modeling performance for a specific task. Remarkably, prompt tuning achieves modeling performance comparable to fine-tuning all layers, yet only necessitates training 0.1\% of the parameters \citep{prefixtuning, ptuning}. On the other hand, adaptation methods involve the insertion of fully connected layers into the transformer blocks \citep{adapter}. These techniques achieve equivalent performance to prompt tuning, albeit slightly more parameter-intensive. \citet{he2022unified} finds an equivalence between prompt tuning and adapter methods: adapter tuning is prompt tuning in series.

Ultimately, the LoRA method has garnered significant popularity \citep{lora}. This technique involves adding a low-rank matrix to certain matrices within the PLLM. The underlying notion is that low-rank matrices encapsulate all the required information for precise task fine-tuning, while PLLM matrices encompass the full spectrum of information from pretraining. Notably, this method is not restricted solely to instruction embedding; it is applicable to a broad array of fine-tuning tasks. Furthermore, when utilized in conjunction with model quantization methods, LoRA extends the capability of fine-tuning numerous PLLMs \citep{qlora}.

\section{Dataset}
Our first objective is to construct a dataset for automatic consolidation. Each sample in this dataset is a triplet of texts \texttt{(existing article, modification section, modified article)} in which the \texttt{modification section} specifies the changes to be made to the \texttt{existing article} to obtain the \texttt{modified article}. Our research shall primarily concentrate on the national consolidation. On a national scope, laws, decrees, and regulations revise existing legal regulations.

\begin{figure}[ht]
    \centering
    \includegraphics[width=\columnwidth]{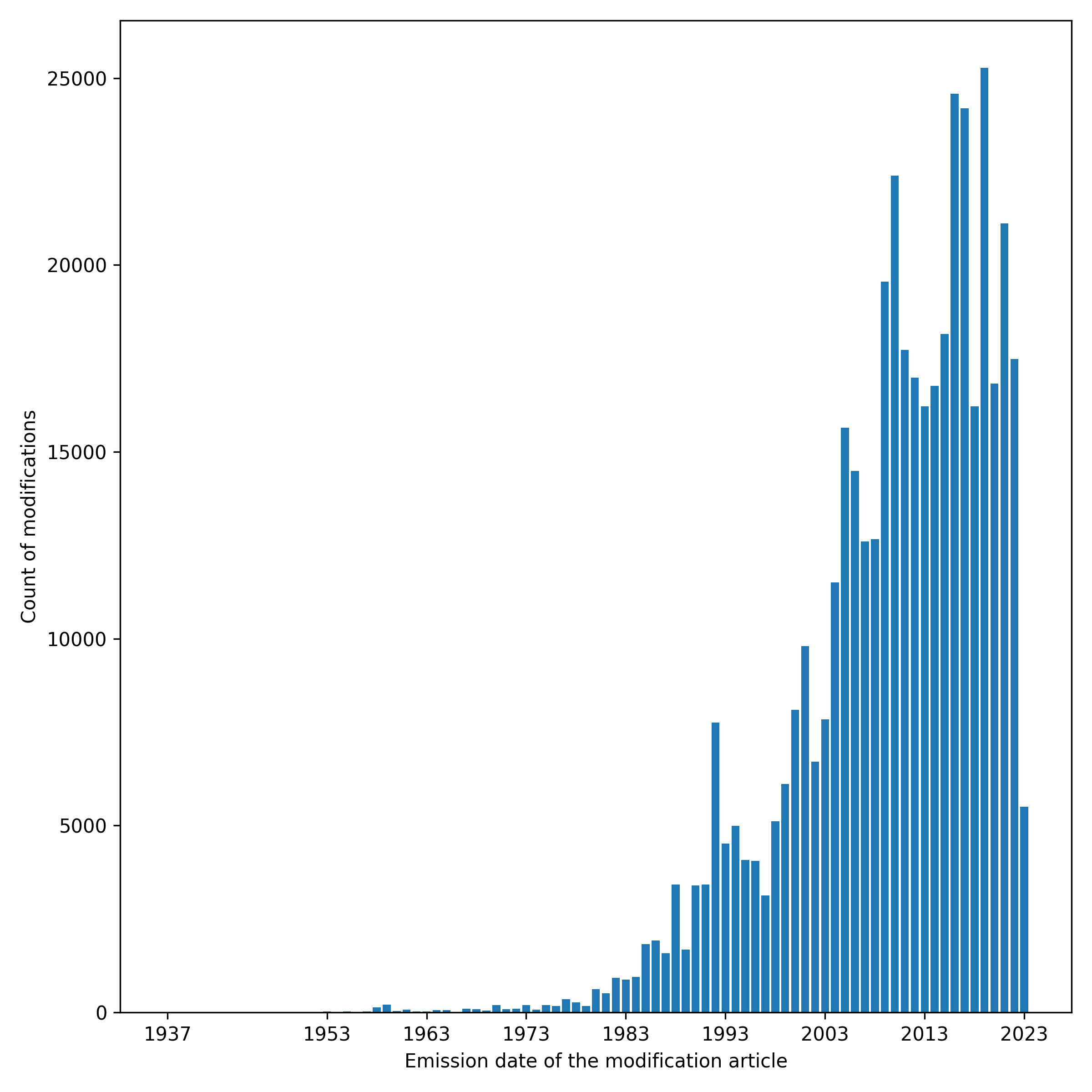}
    \caption{Number of modification sections published per year in France}
    \label{fig:modification_per_year}
\end{figure}

The publication of legal texts has seen significant growth over the past 20 years. Figure \ref{fig:modification_per_year} shows the evolution of the number of modifying articles recorded in France. In 2022, 17487 texts were published from 2512 laws providing modifications on existing laws. We create a dataset of 5000 triplets \texttt{(existing article, modification section, modified article)}. We only keep \texttt{existing articles} that are modified only once. This condition helps avoid \texttt{existing articles} modified by two \texttt{modification sections} simultaneously.

The links between \texttt{(existing article, modification section, modified article)} are publicly available through the Légifrance platform. In the end, we accumulated a dataset comprising 3124 triplets. Example \ref{triplet:main_example} shows a complete sample of the dataset, publicly available on HuggingFace\footnote{\href{https://huggingface.co/datasets/DoctrineAI/legal_consolidation}{Link to dataset}}.

\begin{triplet}[Sample of the dataset]
    \artquote{Existing article}{\textit{Article 10:} \newline Appointments are made each year in the last week of August. The general meeting of the order meets at the courthouse.}

    \artquote{Modification section}{\textit{Article 5:}\newline
        Article 10 is amended as follows: \newline
        1° The words « in the last week of August » are replaced by the words « during the month of December »; \newline
        2° The second sentence is deleted.}

    \artquote{Modified article}{\textit{Article 10:}\newline
        Appointments are made each year during the month of December.
    }\label{triplet:main_example}
\end{triplet}
\label{sec:dataset}

\section{Approaches \& developped methods}

The objective is to modify an existing article by incorporating alterations delineated within a modification document. Initially, a foundational baseline employing a span extraction methodology was developed. Subsequently, this baseline is evaluated against our advanced methodology, which encompasses the fine-tuning of a pre-trained language model.

\subsection{Baseline: span extraction through question answering}

We aim to establish the baseline outlined in Section~\ref{sec:information_extraction}, which involves adapting a BERT model for the question answering task \cite{shi2019simple}. Two distinct models are employed for this purpose. The first model is designed to extract spans that need to be added within the modification section, while the second model identifies spans for deletion within the existing article. Consequently, the span of words identified by the second model can be overwritten by the span generated by the first model. Both approaches utilize the same architecture, employing a CamemBERT Model with a span classification head~\footnote{\href{https://huggingface.co/docs/transformers/en/model_doc/camembert\#transformers.CamembertForQuestionAnswering}{Camembert for question answering}}. This head consists of a linear layer on top of the hidden-state outputs to compute span start logits and span end logits. This model comprises 110M parameters. The batch size is 16 and the learning rate for the Adam optimizer is $2 \times 10^{-5}$. We train for 15 epochs.

\paragraph{Labeling and input format} Example \ref{triplet:span_labels} illustrates the labeling for a sample of the dataset, where spans highlighted in red are predicted by an initial model and subsequently overwritten by spans highlighted in green, as predicted by a second model. This labeling schema facilitates the modeling of three modification types: additions are seamlessly integrated into existing text by substituting blank spaces.
Moreover, we introduced the token \texttt{[NL]} (New Line) prior to the commencement of each paragraph and at the conclusion of each text, as it is denoted that the consolidation process often refers to paragraph. The models acquisition of this token contribute to improved performance in instances exhibiting such patterns.

\begin{triplet}[Labels in the span extraction dataset for a substitution]
    \artquote{Existing article}{[NL]the duties corresponding to the post of Chief State Public Works Engineer in the second group referred to in Article 8 of this Decree are, for the post reporting to the Minister for Foreign Affairs: [NL] \ulcolor[red]{Charged with the duties of Deputy Director of Real Estate Operations in the Real Estate Affairs Department within the General Administration Department}. [NL]}\newline

    \artquote{Modification section}{[NL] The second paragraph of Article 1 of the above-mentioned Order of 4 May 2007 is replaced by the following provisions: [NL] « \ulcolor[green]{Assistant to the Deputy Director of Real Estate Operations.} » [NL]}\newline

    \label{triplet:span_labels}
    \textbf{Legend}: \ulcolor[green]{Span to be predicted by the first model.} \ulcolor[red]{Span to be predicted by the second model to be overwritten by the first span.}
\end{triplet}

When inputting data into the model, the existing article and the modification section are concatenated with a \texttt{[SEP]} token in between. The modification section serves as the "Question" while the existing article acts as the "Paragraph"

\paragraph{Test set and metrics} To assess the model's performance, we test the model on a dataset comprising 302 triplets. Once the spans are predicted, the consolidated text can be reconstructed accordingly. Therefore, it becomes pertinent to utilize a end-to-end oriented metric: word error. Commonly applied in speech-to-text algorithms, the word error measures the number of errors in the transcription of a speech. In our context, this metric assesses the error count within the predicted consolidated text relative to the expected version.

\subsection{Text generation}

Our aim is to leverage generative models to directly predict consolidated texts. Whereas the span extraction method can lead to linguistically nonsensical outcomes in case of prediction errors, generative models ensure the grammatical correctness of generated texts.

\subsubsection{Fune-tuning \& Instruction tuning}

We opt to fine-tune a generative model using the LoRA approach~\citep{qlora}. Given that we are solely focusing on a single task for fine-tuning, it did not seem particularly advantageous to employ a prompt tuning method, which is particularly suited for datasets containing diverse types of instructions. The LoRA  technique was applied to the projection layers of the query, key and value components of the pretrained language model, targeting approximately 3\% of the parameters from the original model.

The prompt format is straightforward and adheres to the conventions commonly employed in instruction tuning. Example~\ref{triplet:prompt_example} illustrates the input format during training. The \texttt{Instruction} corresponds to the modification to be performed, i.e., the modification section. The \texttt{Input} corresponds to the existing article on which the modification is to be applied. Lastly, the expected \texttt{Response} pertains to the modified article. During inference, the \texttt{Response} field is left empty, and the model is tasked with predicting it.

\begin{triplet}[Example of prompt]
    \artquote{\#\#\# Instruction}{\newline Article 10 is amended as follows: \newline
        1° The words « in the last week of August » are replaced by the words « during the month of December »; \newline
        2° The second sentence is deleted.}

    \artquote{\#\#\# Input}{\newline
        Appointments are made each year in the last week of August. The general meeting of the order
        meets at the courthouse.}

    \artquote{\#\#\# Response}{\newline Appointments are made each year during the month of December.
    }\label{triplet:prompt_example}
\end{triplet}

We are employing open-source models that are open for commercial use. Our baseline model is OpenLLama, which is a replication of LLaMa with less intrusive licenses. This model has undergone the same pretraining process as LLaMa and is available in various sizes, ranging from 3 to 13 billion parameters. For training these models, we will utilize Nvidia T4 GPUs with 16GB of memory or Nvidia A10G GPUs with 24GB of memory, depending on the model size.

Consistently across the conducted experiments, certain hyperparameters were kept uniform: the learning rate was set at $3 \times 10^{-4}$, and the LoRA dropout rate was sustained at 5\%. A 4-bit quantization is employed. Only prompts containing fewer than 1024 tokens were selected for use. The micro batch size was determined to be 4, with gradient checkpointing applied after processing every 128 samples. The training duration was limited to 2 epochs.

\subsubsection{Training on the modified article only}

The first experiment involved comparing two models trained with the same prompt, which includes the \texttt{Instruction}, \texttt{Input}, and \texttt{Response} fields. However, the tasks differ: one model is trained to predict the entire prompt (i.e., all three fields: \texttt{Instruction}, \texttt{Input}, and \texttt{Response}), while the other model is trained solely to predict the \texttt{Response} field. In both cases, the full prompt is provided as input during training. Two opposing intuitions were considered. On one hand, training the model to predict the complete prompt could enhance its comprehension of legislative semantics. On the other hand, training the model to predict solely \texttt{Response} field removes certain constraints. For this experiment, we selected two Open-LLaMa models with 3 billion parameters each. The Table \ref{tab:modified_text_only} below summarizes the results. Notably, training a model exclusively on the \texttt{Response} field yields superior performance, of +9.4\%.

\begin{table}[h]
    \centering
    \begin{tabular}{p{25mm}>{\centering\arraybackslash}m{20mm}>{\centering\arraybackslash}m{20mm}}
        \hline
        \textbf{Model trained on} & \textbf{Average Word Error} & \textbf{Median Word Error} \\
        \hline
        Whole prompt              & 18.6                        & 10.5                       \\
        Modified article          & 17.0                        & 7.0                        \\
        \hline
    \end{tabular}
    \caption{\label{tab:modified_text_only}Training on the whole prompt vs. training on the modified article only}
\end{table}

\subsubsection{Influence of cleaning the dataset} We also aimed to examine the influence of dataset quality on consolidation performance. To this end, we selected two OpenLLaMa models with 3 billion and trained them using two distinct consolidation datasets. The second dataset was a cleaned version of the open-sourced dataset, where all consolidation cases that did not involve any modification or involved tables were removed, comprising 1784 triplets.

The results of this comparison highlight the impact of dataset quality on consolidation performance, as shown in Table \ref{tab:dataset_quality}. By using a cleaner dataset that focuses exclusively on meaningful consolidation examples, the model tends to achieve better outcomes, even when compared to a larger dataset that includes less relevant instances. This underscores the significance of dataset quality in influencing model performance for the consolidation task.

\begin{table}[h]
    \centering
    \begin{tabular}{p{25mm}>{\centering\arraybackslash}m{20mm}>{\centering\arraybackslash}m{20mm}}
        \hline
        \textbf{Dataset} & \textbf{Average Word Error} & \textbf{Median Word Error} \\
        \hline
        Full dataset     & 17.0                        & 7.0                        \\
        \hline
        Curated dataset  & 12.0                        & 4.0                        \\
        \hline
    \end{tabular}
    \caption{Influence of the quality of the training dataset}\label{tab:dataset_quality}
\end{table}

\subsubsection{Influence of the size of the low rank matrix} The LoRA finetuning method encompasses two hyperparameters: the rank $r$ of the added matrices and the multiplier $\alpha$. The multiplier $\alpha$ operates as a learning rate for the added matrices and exhibits relatively modest effects once it reaches a sufficiently high value. It was set as twice the value of $r$. The matrix rank $r$ significantly impacts the model's performance. A smaller $r$ suggests limited fine-tuning, where the model requires minimal adaptation to accomplish the intended task. In contrast, a larger $r$ implies extensive retraining, almost akin to starting from scratch. Table \ref{tab:rank_value} showcases the performance of two models, each utilizing different $r$ values. Notably, the model with the higher $r$ value attains slightly better consolidation capabilities. It can be observed that the model with a higher rank value $r$ trains faster but eventually converges to a similar value as the other model.

\begin{table}[ht]
    \centering
    \begin{tabular}{>{\centering\arraybackslash}m{20mm}>{\centering\arraybackslash}m{20mm}>{\centering\arraybackslash}m{20mm}}
        \hline
        \textbf{Rank $r$} & \textbf{Average Word Error} & \textbf{Median Word Error} \\
        \hline
        16                & 12.0                        & 4.0                        \\
        \hline
        64                & 11.7                        & 4.0                        \\
        \hline
    \end{tabular}
    \caption{\label{tab:rank_value}Influence of the rank of the added matrices}
\end{table}

\subsubsection{Influence of the size of the PLLM} We also examined the impact of the PLLM size on consolidation performance. To do so, we compared three OpenLLaMa models with 3 billion, 7 billion, and 13 billion parameters, respectively trained on a curated dataset with a large low-rank $r$. Despite being more challenging to fine-tune, larger models generally exhibit better performance due to their increased information retention capacity. We further compared these models with a 13-billion-parameter OpenLLaMa model that had already undergone an initial round of fine-tuning on an instruction dataset.

Table \ref{tab:size_PLLM} outlines the results. It's observed that, on average, the number of errors is lower for the 3-billion-parameter model compared to the 7-billion-parameter model. However, the number of errors is higher in terms of median values. The 7-billion-parameter model generally predicts better modified articles, but in some cases, the modified article is significantly worse from the expected text. This can be attributed to the fact that the 7-billion-parameter model possesses a larger generative capacity. As a result, in complex consolidation examples, the 7-billion-parameter model might "hallucinate" and generate interpretations of the texts, whereas the 3-billion-parameter model tends to generate the un-consolidated existing article. This effect disappear in the 13-billion-parameter model which showcases a considerable performance gap. The 13-billion-parameter model, which was pre-fine-tuned on an instruction dataset, further enhances consolidation performance.

\begin{table}[h]
    \centering
    \begin{tabular}{p{27mm}>{\centering\arraybackslash}m{20mm}>{\centering\arraybackslash}m{20mm}}
        \hline
        \textbf{Model size} & \textbf{Average Word Error} & \textbf{Median Word Error} \\
        \hline
        3b                  & 10.0                        & 2.0                        \\
        7b                  & 13.5                        & 0.5                        \\
        13b                 & 6.07                        & 0                          \\
        13b pre-finetuned   & 5.09                        & 0                          \\
        \hline
    \end{tabular}
    \caption{\label{tab:size_PLLM}Influence of the size of the PLLM}
\end{table}

\label{sec:finetuning}
\subsection{Comparing all methods}

We now proceed to compare the different approaches employed: span extraction and the generative method, on a more challenging dataset. While it would have been ideal to maintain the same test set as used in previous sections, it became evident that the models were insufficiently distinguishable based on that test set alone. Therefore, we opted to construct a second, more complex test set, utilizing the legal provisions from the previous year's PLF. Additionally, we compare our results with OpenAI's GPT-3.5 and GPT-4 models to further contextualize model performance.

\begin{table}[h]
    \centering
    \begin{threeparttable}
        \begin{tabular}{p{35mm}>{\centering\arraybackslash}m{10mm}>{\centering\arraybackslash}m{22mm}}
            \hline
            \textbf{Approach}            & \textbf{Prompt size} & \textbf{Average Word Error (95\% CI)} \\
            \hline
            Span Extraction              & 512                  & 36.2\tnote{*} (31.4)                  \\
            Generative models            &                      &                                       \\
            \quad Open-LLaMA 3b          & 1024                 & 65.5 (29.2)                           \\
            \quad Open-LLaMA 13b         & 1024                 & \textbf{20.7} (7.79)                  \\
            \hline
            \quad      GPT3.5-turbo-0613 & 4k                   & 44.8 (27.5)                           \\
            \quad      GPT4-0613         & 8k                   & \textbf{9.41} (3.58)                  \\
            \hline
        \end{tabular}
        \begin{tablenotes}\footnotesize
            \item[*] Computed only on single modifications
        \end{tablenotes}
    \end{threeparttable}
    \caption{\label{tab:comparison}Comparison of the proposed approach with the baseline}
\end{table}

\begin{figure*}[!htb]
    \centering
    \includegraphics[width=\linewidth]{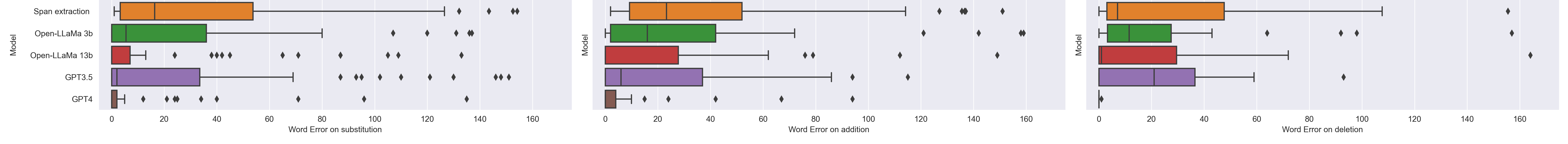}
    \caption{Word error distributions per model per modification type}
    \label{fig:model_comparison_per_modification}
\end{figure*}

The results are denoted in Table \ref{tab:comparison}. We first observe that the generative models yield the best performances. While these models generally produce highly accurate consolidations, in certain cases, the consolidation can result in aberrations, leading to hallucinations and the generation of lengthy texts, resulting in substantial consolidation errors. The distributions of word errors for each model and error type (addition, deletion, substitution) are depicted in Figure \ref{fig:model_comparison_per_modification}.  Additionally, it is notable that GPT4 demonstrates superior performance, while our best model (Open-LLaMa 13b) falls between GPT3.5 and GPT4.

\section{Application on a bill}

Encouraged by our model's promising performance, we embarked on live automatic consolidation of the \textit{Projet de Loi de Finance 2024} from September 2023 to December 2023.
This bill was proposed on 26th of September, 2023 and contained 60 articles. After multiples debates at the parliements, the bill was promulgated on 29th of December, 2023 with 264 articles. This is a highly complex bill.

\subsection{Pipeline}

\begin{figure*}[htb]
    \centering
    \includegraphics[width=\linewidth]{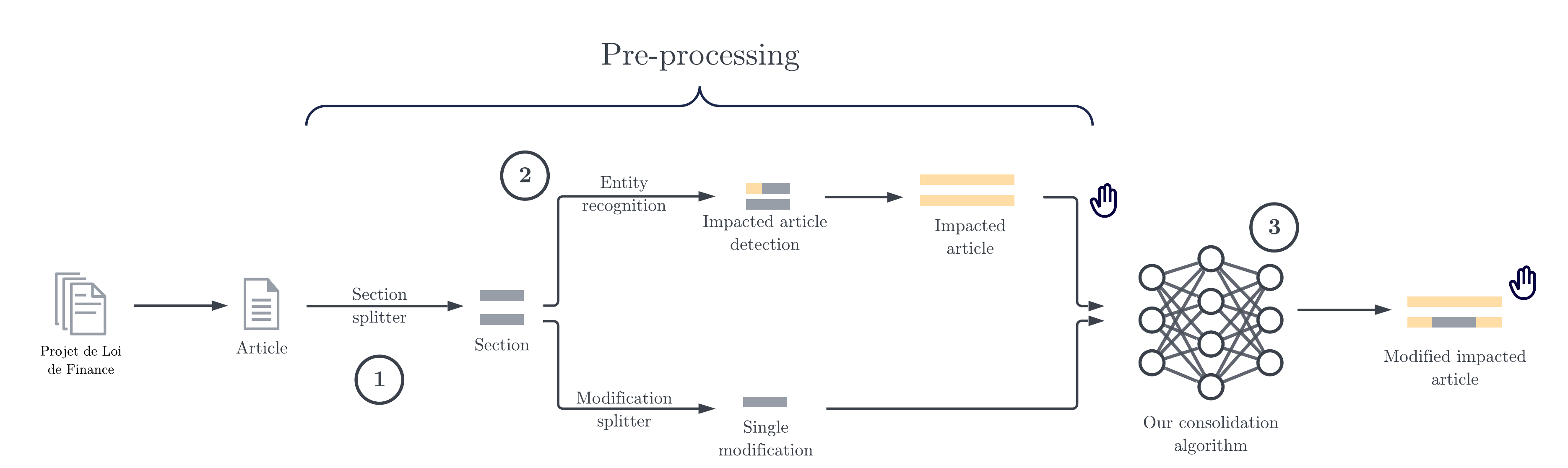}
    \caption{Full consolidation pipeline}\label{fig:consolidation_pipeline}
\end{figure*}

Our consolidation pipeline is depicted in Figure \ref{fig:consolidation_pipeline}. This pipeline was up during the four months of life of the bill. The pre-processing consists of three primary steps.

\subsubsection{Section splitter}
A bill is structured into articles, each specifying modifications to current laws on particular topics. Consequently, a bill's article might introduce several changes to numerous laws. A section division using regular expressions is therefore created to break down the bill's article into these distinct sections.  This splitter leverages the hierarchical structure of the bill's article to efficiently segment it into components.

\subsubsection{Entity recognition}
We employ an already fine-tuned entity recognition model  system to identify the specific law articles targeted by each section. Upon identifying these articles, we retrieve their contents for further processing.

\subsubsection{Our consolidation algorithm}
We use our best model to generate the consolidated text.

\subsection{Results}

In this section, we delineate the consolidation process undertaken as of 16th of December, 2023. At this juncture, the legislative bill comprised 271 articles. Upon division, this legislative text was found to encompass 1399 simple modifications applicable to 606 articles of law.

Our pipeline incorporates two instances of human intervention, symbolized by hand icons, primarily focused on verification rather than labeling. The law article detection phase, leveraging an existing entity recognition component, achieved an 82.0\% success rate.
To quantify the success rate of our algorithm, we executed the consolidation process on the legislative bill using both GPT-4 and our best model, OpenLLaMa-13, generating two sets of predictions. Subsequently, we scrutinized and amended the predictions made by GPT-4 to produce a third set, representing human annotations. For a prediction and an annotation, we removed special characters from each string, such as accents, commas, and line breaks, to facilitate the comparison of the raw texts. However, it exists two cases where the consolidation process can't be done: the presence of tables and lengthy prompts.
Table~\ref{tab:correct_consolidation_rate} presents the rate of possible consolidations along the rate of correct consolidation for both algorithms.

\begin{table}[htb]
    \centering
    \begin{tabular}{p{25mm}>{\centering\arraybackslash}m{20mm}>{\centering\arraybackslash}m{20mm}}
        \hline
        \textbf{Model}                           &
        \textbf{Rate of possible consolidations} & \textbf{Correctness rate among possible consolidations}          \\
        \hline
        Our model                                & 49.8\%                                                  & 63.2\% \\
        GPT4-0613                                & 91.3\%                                                  & 61.4\% \\
        \hline
    \end{tabular}
    \caption{\label{tab:correct_consolidation_rate}Correct consolidation rate}
\end{table}

Our Open-LLaMa-13b model faces challenges due to its limited context size, allowing application in only 49.8\% of consolidation cases. Conversely, GPT4-0613 encounters difficulties in consolidating only 8.7\% of cases, all related to the inclusion of tables. In terms of correctness rates, both models achieve 63.2\% and 61.4\% respectively, considering their respective possible consolidations. While our algorithm appears to achieve a higher correctness rate, it's crucial to note that it consolidates far fewer samples with much smaller prompt sizes compared to GPT4, which consolidates most of them.

\begin{figure}[htb]
    \centering
    \includegraphics[width=\linewidth]{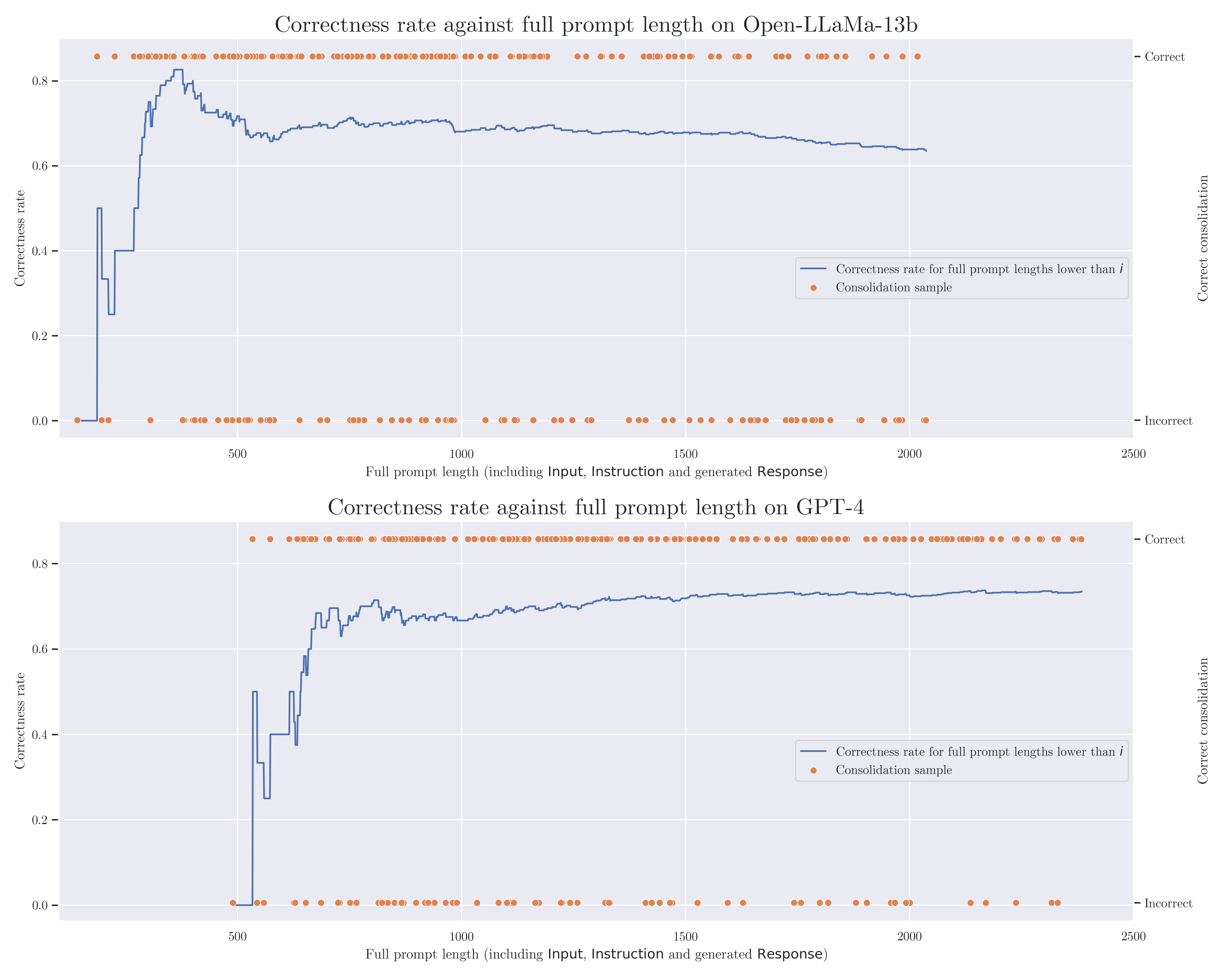}
    \caption{The correctness rates against prompt length are plotted for Open-LLaMa-13b and GPT-4 on the same consolidation samples (49.8\% of the PLF). Each dot represents a sample of the PLF consolidation, indicating whether it is correct or not. The curve at prompt length $i$ illustrates the rate of correct consolidation among samples with a prompt length less than $i$.}\label{fig:correctness_rate_against_length}
\end{figure}

In Figure \ref{fig:correctness_rate_against_length}, we depict the correctness rate against the full prompt length, including the generated \texttt{Response}, for both models in cases possible for our Open-LLaMa-13b. Here, GPT4-0613 achieves a 73.6\% correctness rate. Notably, the full prompt length for the GPT-4 model slightly differs due to the inclusion of few-shot examples. Both models exhibit differing behaviors in correctness rates against full prompt length. Open-LLaMa-13b peaks for full prompt lengths below 1000 tokens, with performance gradually decreasing for larger prompts, highlighting attention mechanism limitations. Conversely, GPT4-0613 demonstrates consistent performance across varying prompt lengths, showing no impact from larger prompts.

\section{Conclusion}
This research implements a generative method to automate legislative text consolidation, demonstrating a significant capability to process and automatically apply changes to legislative texts. We determined that the quality of the dataset and the size of the pre-trained model were two parameters that most significantly influenced consolidation performance.  Despite exceptional performances of GPT4,  in the end, we ideally prefer to use an open-source model for handling legal data due to its sensitivity. The consolidation, led on a real-time legislative bill, proved to be highly effective, although occasional issues in the generation process could result in nonsensical consolidations.

Moving forward, our objective is to delve into advanced fine-tuning strategies and broaden our methodology to encompass additional models. On one side, there exists a variety of models equipped with commercial licenses, such as LLaMA 3.1, that offer new possibilities for exploration. These models often feature larger context windows, enabling the consolidation of more samples. On the other side, innovative fine-tuning techniques are being developed, such as the Mixture of LoRA Experts approach. This technique is designed to fine-tune each expert within a Mixture of Experts.

This research opens promising avenues for integrating generative methods into legal processes, with the hope of radically transforming legal practice.

\bibliography{custom}



\end{document}